\documentclass[USletter,10pt,english]{article}
\usepackage[margin=0.8in]{geometry}
\usepackage{graphicx}      
\usepackage[font=normalsize]{caption}
\usepackage{amsfonts}
\usepackage{amsmath}
\usepackage{amssymb}
\usepackage{bm}
\usepackage{algorithm}
\usepackage{algpseudocode}
\usepackage{lipsum}
\usepackage{color}
\usepackage[breaklinks]{hyperref}
\usepackage{mathrsfs}
\usepackage{psfrag,epsf}
\usepackage{amsthm}
\usepackage{url}

\usepackage{breakurl}
\usepackage{float}

\DeclareMathOperator*{\argmax}{arg\,max}

\newif\ifarxiv
\arxivtrue
\arxivfalse

\usepackage{amsbsy}
\usepackage{mathrsfs}
\usepackage{mathtools}
\usepackage{setspace,multirow, url,  subfig, bm, titlesec, textcomp, gensymb}

\def \exp {\mathrm{exp}}

\def \vec {\mathrm{vec}}

\newcommand{\Mnorm}[2]{{\left\vert\kern-0.30ex\left\vert\kern-0.30ex\left\vert #1 
		\right\vert\kern-0.30ex\right\vert\kern-0.30ex\right\vert}}
\newcommand{\Opnorm}[3]{{\left\vert\kern-0.25ex\left\vert\kern-0.25ex\left\vert #1 
		\right\vert\kern-0.25ex\right\vert\kern-0.25ex\right\vert}_{#2 \to #3}}
\newcommand{\norm}[2]{{\left\vert\kern-0.30ex\left\vert #1 
		\right\vert\kern-0.30ex\right\vert}}

\newcommand{\innerproductminconstant}[1]{\psi_0}

\newtheorem{theorem}{Theorem}

\newtheorem{lemma}{Lemma}

\newif\ifarxiv
\arxivtrue
\arxivfalse

\begin{document}
\ifarxiv
\doublespacing
\onecolumn
\else \fi

\title{Worst-case Performance of Greedy Policies in Bandits with Imperfect Context Observations} 
\author{Hongju Park and Mohamad Kazem Shirani Faradonbeh}%
\date{}
\maketitle


\begin{abstract}  
Contextual bandits are canonical models for sequential decision-making under uncertainty in environments with time-varying components. In this setting, the expected reward of each bandit arm consists of the inner product of an unknown parameter with the context vector of that arm. The classical bandit settings heavily rely on assuming that the contexts are fully observed, while study of the richer model of imperfectly observed contextual bandits is immature. This work considers Greedy reinforcement learning policies that take actions as if the current estimates of the parameter and of the unobserved contexts coincide with the corresponding true values. We establish that the non-asymptotic worst-case regret grows  poly-logarithmically with the time horizon and the failure probability, while it scales linearly with the number of arms. Numerical analysis showcasing the above efficiency of Greedy policies is also provided.
\end{abstract}



\section{Introduction}
\label{sec:1}

Contextual bandits are ubiquitous models sequential decision making in environments with finite action spaces. The range of applications is extensive and includes different problems that time-varying and action-dependent information are important, such as personalized
recommendation of news articles, healthcare interventions, advertisements, and clinical trials~\cite{li2010contextual,bouneffouf2012contextual,tewari2017ads,nahum2018just,durand2018contextual,varatharajah2018contextual,ren2020dynamic}. 

In many applications, consequential variables for decision making are not perfectly observed. Technically, the context vectors are often observed in a partial, transformed, and/or noisy manner~\cite{bensoussan2004stochastic,bouneffouf2017context,tennenholtz2021bandits}. In general, sequential decision making algorithms under imperfect observations provide a richer class of models compared to those of perfect observations. Accordingly, they are commonly used in different problems, including space-state models for robot control and filtering \cite{roesser1975discrete,nagrath2006control,kalman1960new,stratonovich1960application}. 

We study contextual bandits with imperfectly observed context vectors. The probabilistic structure of the problem under study as time proceeds, is as follows. At every time step, there are $N$ available actions (also referred to as `arms'), and the unobserved context of arm $i$ at time $t$, denoted by $x_i(t) \in \mathbb{R}^{d_x}$, is generated according to a multivariate normal distribution $\mathcal{N}(0_{d_x}, \Sigma_x)$. Moreover, the corresponding observation (i.e., output) is $y_i(t)\in \mathbb{R}^{d_y}$, while the stochastic reward $r_i(t)$ of arm $i$ is determined by the context and the unknown parameter $\mu_*$. Formally, we have
\begin{flalign}
y_i(t) &= A x_i(t) + \zeta_{i}(t)\label{eq:om},\\
r_i(t) &= x_i(t)^\top \mu_* + \psi_{i}(t),\label{eq:reward}
\end{flalign}
where $\zeta_{i}(t)$ and $\psi_{i}(t)$ are the noises of observation and reward, which are identically distributed and independent following the distributions $\mathcal{N}(0_{d_y}, \Sigma_{y})$ and  $\mathcal{N}(0, \gamma_r^2)$, respectively. Further, the $d_y \times d_x$ sensing matrix $A$ captures the relationship between $x_i(t)$ and the noiseless portion of $y_i(t)$. The above structure holds for all arm $i$ and time $t$. From a dynamical system point of view, the setting can be understood as memoryless dynamical systems.

At each time, the goal is to learn to choose the optimal arm $a^*(t)$ maximizing the reward, by utilizing the available information by time $t$. That is, the agent chooses an arm based on the data collected so far from the model in \eqref{eq:om}; $\{y_{i}(t)\}_{1\leq i \leq N},\{a(\tau)\}_{1\leq \tau \leq t-1},\{y_{a(\tau)}(t)\}_{1\leq \tau \leq t-1},\{r_{a(\tau)}\}_{1\leq \tau \leq t-1}$. So, the resulting reward will be provided to the agent according to the equation in \eqref{eq:reward}. Clearly, to choose high-reward arms, the agent needs accurate estimates of the unknown parameter $\mu_*$, as well as those of the contexts $x_i(t)$, for $i=1, \cdots, N$. However, because $x_i(t)$ is not observed, the estimation of $\mu_*$ is available only through the output $y_i(t)$. Thereby, design of efficient reinforcement learning algorithms with guaranteed performance is challenging. 

Bandits are thoroughly investigated in the literature, assuming that $\{x_i(t)\}_{1\leq i \leq N}$ are perfectly observed. Early papers focus on the method of Upper-Confident-Bounds (UCB) for addressing the exploitation-exploration trade-off~\cite{lai1985asymptotically,abe1999associative,auer2002using,abbasi2011improved,chu2011contextual}. UCB-based methods take actions following optimistic estimations of the parameters, and are commonly studied in reinforcement learning \cite{abbasi2011regret,faradonbeh2020optimism}. Another popular and efficient family of policies use randomized exploration, usually in the Bayesian form of Thompson sampling~\cite{chapelle2011empirical, agrawal2013thompson,faradonbeh2019applications,faradonbeh2020input,faradonbeh2020adaptive,modi2020no}. For contextual bandits that contexts are generated under certain conditions, exploration-free policies with Greedy nature can expose efficient performance~\cite{bastani2021mostly}.

Currently, theoretical results for bandits with imperfect context observations are incomplete. For contextual bandits with noisy  observations with the same dimension as that of contexts, asymptotic analyses are available for a UCB-type algorithms \cite{yun2017contextual}, and Thompson sampling \cite{park2021analysis}, or in presence of additional information~\cite{tennenholtz2021bandits}. Moreover, the relationship between the regret and gained information under the uncertainty of observations are analyzed~\cite{lattimore2021mirror, lattimore2022minimax}. However, analyses about contextual bandits with noisy transformed observations, whose dimension can be different from that of contexts, are scarce.
Lastly, numerical analysis shows that Greedy algorithms outperform Thompson sampling under imperfect context observations in the suggested framework \cite{park2022efficient}. Therefore, this work focuses on the non-asymptotic theoretical analysis of Greedy policies for imperfectly observed contextual bandits. 


However, comprehensive analyses and non-asymptotic theoretical performance guarantees for general output observations are not currently available and are adopted as the focus of this work. We perform the \emph{finite-time worst-case} analysis of Greedy reinforcement learning algorithms for imperfectly observed contextual bandits. We establish efficiency and provide high probability upper bounds for the regret that consists of poly-logarithmic factors of the time horizon and of the failure probability. Furthermore, the effects of other problem parameters such as the number of arms and the dimension are fully characterized. Illustrative numerical experiments showcasing the efficiency are also provided.

To study the performance of reinforcement learning policies, different technical difficulties arise in the high probability analyses. First, one needs to study the eigenvalues of the empirical covariance matrices, since the estimation accuracy depends on them. Furthermore, it is required to consider the number of times the algorithm selects sub-optimal arms. Note that both quantities are stochastic and so worst-case (i.e., high probability) results are needed for a statistically dependent sequence of random objects. To obtain the presented theoretical results, we employ advanced technical tools from martingale theory and random matrices. Indeed, by utilizing concentration inequalities for matrices with martingale difference structures, we carefully characterise the effects of order statistics and tail-properties of the estimation errors. 

The remainder of this paper is organized as follows. In Section~\ref{sec:2}, we formulate the problem and discuss the relevant preliminary materials. Next, a Greedy reinforcement learning algorithm for contextual bandits with imperfect context observations is presented in Section \ref{sec:3}. In Section \ref{sec:4}, we provide theoretical performance guarantees for the proposed algorithm, followed by numerical experiments in Section~\ref{sec:5}. Finally, we conclude the paper and discuss future directions in Section~\ref{sec:6}.

We use $A^\top$ to refer to the transpose of the matrix $A \in \mathbb{C}^{p \times q}$. For a vector $v \in \mathbb{C}^d$, we denote the $\ell_2$ norm by $\|v\| = \left(\sum_{i=1}^d |v_i|^2\right)^{1/2}$. Additionally, $C(A)$ and $C(A)^\perp$ are employed to denote the column-space of the matrix $A$ and its orthogonal subspace, respectively. Further, $P_{C(A)}$ is the projection operator onto $C(A)$. Moreover, $\lambda_{\min}(A)$ and $\lambda_{\max}(A)$ denote the minimum and maximum eigenvalues of the symmetric matrix $A$, respectively. Finally, $O(\cdot)$ denotes the order of magnitude, $\{X_i\}_{i \in E} = \{X_i: i \in E\}$ and $I(\cdot)$ is the indicator function. 

\section{Problem Formulation} \label{sec:2}

First, we formally discuss the problem of contextual bandits with imperfect context observations. A bandit machine has $N$ arms, each of which has its own unobserved context $x_i(t)$, for $i \in \{1,\cdots,N\}$. Equation \eqref{eq:om} presents the observation model, where the observations $\{y_i(t)\}_{1\leq i \leq N}$ are linearly transformed functions of the contexts, perturbed by additive noise vectors $\{\zeta_{i}(t)\}_{1\leq i \leq N}$. Equation \eqref{eq:reward} describes the process of reward generation for different arms, depicting that \emph{if} the agent selects arm $i$, then the resulting reward is an \emph{unknown} linear function of the unobserved context vector, subject to some additional randomness due to the reward noise $\psi_i(t)$.

The agent aims to maximize the cumulative reward over time, by utilizing the sequence of observations. To gain the maximum possible reward, the agent needs to learn the relationship between the rewards $r_i(t)$ and the observations $y_i(t)$. For that purpose, we proceed by considering the conditional distribution of the reward $r_i(t)$ given the observation $y_i(t)$, i.e., $\mathbb{P}(r_i(t)|y_i(t))$, which is
\begin{flalign}
 \mathcal{N}(y_i(t)^\top D^\top \mu_*, \gamma_{ry}^2),\label{eq:condr}
\end{flalign}
where $D=(A^\top \Sigma^{-1}_y A + \Sigma^{-1}_x)^{-1} A^\top \Sigma^{-1}_y$ and $\gamma_{ry}^2 = \mu_*^\top(A^{\top} \Sigma_y^{-1} A + \Sigma_x^{-1})^{-1}  \mu_* + \gamma_r^2$.

Based on the conditional distribution in \eqref{eq:condr}, in order to maximize the expected reward given the observation, we consider the conditional expectation of the reward given the observations, $y_i(t)^\top D^\top \mu_*$. So, letting $\eta_* = D^\top \mu_*$ be the transformed parameter, we focus on the estimation of $\eta_*$. The rationale is twofold; first, the conditional expected reward can be inferred with only knowing $\eta_*$, regardless of the exact value of  the true parameter $\mu_*$. Second, $\mu_*$ is not estimable when the rank of the sensing matrix $A$ in the observation model is less than the dimension of $\mu_*$. Indeed, estimability of $\mu_*$ needs the restrictive assumptions of non-singular $A$ and $d_y \geq d_x$. 

The optimal policy that reinforcement learning policies need to compete against knows the true parameter $\mu_*$. That is, to maximize the reward given the output observations, the optimal arm at time $t$, denoted by $a^*(t)$, is
\begin{flalign}
a^*(t) = \argmax_{1\leq i \leq N} y_i(t)^\top \eta_*.\label{eq:op}
\end{flalign}
Then, the performance degradation due to uncertainty about the environment that the parameter $\mu_*$ represents, is the assessment criteria for reinforcement learning policies. So, we consider the following performance measure, which is commonly used in the literature, and is known as \emph{regret} of the reinforcement learning policy that selects the sequence of actions $a(t), t =1,2, \cdots$:
\begin{flalign}
\mathrm{Regret}(T) &= \sum_{t=1}^T \left(y_{a^*(t)}(t) - y_{a(t)}(t)\right)^\top \eta_*.\label{eq:reg}
\end{flalign}
In other words, the regret at time $T$ is the total difference in the obtained rewards, up to time $T$, where the difference at time $t$ is between the optimal arms $a^*(t)$ and the arm $a(t)$ chosen by the reinforcement learning policy based on the output observations by the time $t$. Note that this difference does not depends on the unknown contexts $\{x_i(t)\}_{1\leq i \leq N}$. That is, the arm maximizing $x_i(t)^\top \mu_*$ is not guaranteed to be $a^*(t)$, since $y_i(t)^\top \eta_*$ is a realized value of a random variable centered at $x_i(t)^\top \mu_*$.
\section{Reinforcement Learning Policy}
\label{sec:3}

In this section, we explain the details of the Greedy algorithm for contextual bandits with imperfect observations. Although inefficient in some reinforcement learning problems, Greedy algorithms are known to be efficient under certain conditions such as covariate diversity~\cite{bastani2021mostly}. Intuitively, the latter condition expresses that the context vectors cover all directions in $\mathbb{R}^{d_x}$ with a non-trivial probability, so that additional exploration is not necessary. 

As discussed in Section \ref{sec:2}, it suffices for the policy to learn to maximize 
\begin{eqnarray}
\mathbb{E}[r_i(t)|y_i(t)]=y_i(t)^\top \eta_*.
\end{eqnarray}
To estimate the quantity $y_i(t)^\top \eta_*$, we use the least-squares estimate 
\begin{eqnarray}
\widehat{\eta}(t) = \argmax_{\eta} \sum_{\tau=1}^t (r_{a(\tau)}(\tau) - y_{a(\tau)}(\tau)^\top \eta)^2,\label{eq:lse}
\end{eqnarray}
in lieu of the truth $\eta_*$. So, the Greedy algorithm selects the arm $a(t)$ at time $t$, such that
 \begin{eqnarray}
 a(t) = \argmax_{1\leq i \leq N} y_i(t)^\top \widehat{\eta}(t).\label{eq:pol}
 \end{eqnarray}
The recursions to update the parameter estimate $\widehat{\eta}(t)$ and the empirical inverse covariance matrix $B(t)$ based on \eqref{eq:lse} are as follows:
\begin{eqnarray}
B(t+1) &=& B(t) + y_{a(t)}(t)y_{a(t)}(t)^\top\label{eq:B}\\
\widehat{\eta}(t+1) &=& B(t+1)^{-1} \left(B(t) \widehat{\eta}(t) + y_{a(t)}(t) r_{a(t)}(t)\right),~~~~~\label{eq:etahat}
\end{eqnarray}
where the initial values consist of $B(1) = \Sigma^{-1}$, for some arbitrary postitive definite matrix $\Sigma$, and $\widehat{\eta}(1)= \eta$ for an arbitrary vector $\eta$ in $\mathbb{R}^{d_y}$. Algorithm~\ref{algo} describes the pseudo-code for the Greedy algorithm.

\begin{algorithm}[h] 
	\begin{algorithmic}[1]
		\State Set $B(1) = \Sigma^{-1}$, $\widehat{\eta}(1) = \eta$
		\For{$t = 1,2, \dots, $} 
		\State Select arm $a(t)= \underset{1 \leq i \leq N}{\arg\max}~ y_i(t)^{\top} \widehat{\eta}(t)$
		\State Gain reward $r_{a(t)}(t) = x_{a(t)}(t)^\top \mu_* + \psi_{a(t)}(t)$
		\State Update $B(t+1)$ and $\widehat{\eta}(t+1)$ by \eqref{eq:B} and  \eqref{eq:etahat}
		\EndFor
	\end{algorithmic}
\caption{: Greedy policy for contextual bandits with imperfect context observations}  \label{algo}
\end{algorithm}

\section{Theoretical Performance Guarantees}
\label{sec:4}

In this section, we present a theoretical result for Algorithm \ref{algo} presented in the previous section. The result provides a worst-case analysis and establishes a high probability upper-bound for the regret in \eqref{eq:reg}.

\begin{theorem}
Assume that Algorithm \ref{algo} is used in a contextual bandit with $N$ arms and the output dimension $d_y$. Then, with probability at least $1-4\delta$, we have 
\begin{flalign*}
    &\mathrm{Regret}(T) = \nonumber\\
    &O\left( \frac{(\lambda_{a2}+\lambda_{y2})\gamma_{ry}}{\lambda_{a1}+\lambda_{y1}} Nd_y^{3/2} \left(\log \frac{Nd_y T}{\delta}\right)^{5/2} \log \frac{d_y T}{\delta}  \right),
\end{flalign*}
where $\lambda_{a1} = \lambda_{\min}(A\Sigma_xA^\top)$, $\lambda_{a2} = \lambda_{\max}(A\Sigma_xA^\top)$, $\lambda_{y1}=\lambda_{\min}(\Sigma_y)$, $\lambda_{y2}=\lambda_{\max}(\Sigma_y)$ and $\gamma_{ry}^2$ is the conditional variance in \eqref{eq:condr}.
\end{theorem}
The regret bound above scales linearly with the number of arms $N$, with $d_y^{3/2}$ for the dimension of the observations $d_y$, and poly-logarithmically with time $T$. The dimension of unobserved context vectors does not affect the regret because the optimal policy in \eqref{eq:op} does not have the exact values of the context vectors. So, similar to the reinforcement learning policy, the optimal policy needs to estimate the contexts as well, as $y_i(t)^\top \eta_*$ in \eqref{eq:op} is an estimate of $x_i(t)^\top \mu_*$ for the optimal policy to find the optimal arm.

The rationale of the linear growth of the regret with $N$ is that a policy is more likely to choose one of sub-optimal arms, when more sub-optimal arms exist, incurring more regret. In addition, the quadratic term of $d_y$ and the maximum eigenvalue $\lambda_{a2}$ are generated by the use of truncation for the $\ell_2$ norm of vector $\left(\|y_i(t)\|_2^2 = O(\lambda_{a2}d_y v_T(\delta)^2\right))$ as well as the matrix Azuma's inequality. 
Further, the poly-logarithmic terms of $T$, $N$, $d_y$ and $\delta$, $\left(\log (Nd_yT/\delta)\right)^{5/2}\log (d_yT/\delta)$, are originated in the truncation event and the Azuma's inequality. 
Lastly, the minimum eigenvalue $\lambda_{a1}$ and the conditional reward variance $\gamma_{ry}^2$ are associated with the variance of the estimator $\widehat{\eta}(t)$, whose larger value causes a greater regret.
\begin{proof}
We use the following intermediate results, whose proofs are delegated to Appendices. For simplicity, let $\widehat{\eta}(1)$ be a random variable with $\mathbb{E}[\widehat{\eta}(1)] = \eta_*$ and $\mathrm{Cov}(\widehat{\eta}(1)) = \Sigma^{-1}\gamma_{ry}^2$ so that $\mathbb{E}[\widehat{\eta}(t)] = \eta_*$ and $\mathrm{Cov}(\widehat{\eta}(t)|B(t)) = B(t)^{-1}\gamma_{ry}^2$ for all $t$. First, for $T>0$ and $0<\delta<0.25$, we define
\begin{eqnarray}
W_T = \left\{\underset{\{1\leq \tau \leq t~and~1\leq i \leq N\}}{\max} ||S_y^{-1/2}y_i(\tau)||_{\infty} \leq  v_T(\delta)\right\} \label{eq:WT},
\end{eqnarray} 
where $v_T(\delta) = (2\log (Nd_yT/\delta) )^{1/2}$.
\begin{lemma}
For the event $W_T$ defined in \eqref{eq:WT}, we have $\mathbb{P}(W_T) \geq 1 -\delta$.\label{lem:1}
\end{lemma}
Lemma \ref{lem:1} guarantees that all the observation up to time $T$ are generated in the truncation event $W_T$ with the probability at least $1-\delta$. 
\begin{lemma}
Let $\sigma\{X_1,\dots,X_n\}$ be the sigma-field generated by random vectors $X_1,\dots,X_n$. For the observation of chosen arm $y_{a(t)}(t)$ at time t, the estimator $\widehat{\eta}(t)$ defined in \eqref{eq:etahat}, and the filtration $\{\mathscr{F}_{t}\}_{1\leq t \leq T}$ defined according to  $$\mathscr{F}_{t} = \sigma\{\{a(\tau)\}_{1\leq \tau \leq t},\{y_i(\tau)\}_{1\leq \tau \leq t, 1\leq i \leq N}, \{r_{a(\tau)}(\tau)\}_{1\leq \tau \leq t}\},$$
we have 
\begin{flalign}
\mathbb{E}[ V_t|\mathscr{F}_{t-1}]
= P_{C(S_y^{1/2}\widehat{\eta}(t))} (k_N-1)  + I_{d_y},\nonumber
\end{flalign}
where $V_t=S_y^{-1/2} y_{a(t)}(t)y_{a(t)}(t)^{\top} S_y^{-1/2}$ and $k_N = \mathbb{E}\left[\left(\underset{1 \leq i \leq N}{\max}\{Z_i\}\right)^2\right]$ for $N$ independent $Z_i$ with the standard normal distribution and $S_y = \mathrm{Cov}(y_i(t))$. That is, $k_N$ is the expected maximum of $N$ independent standard normal random variables.
\label{lem:2}
\end{lemma}
Lemma \ref{lem:2} sets the stage for analysis of the (unnormalized) empirical inverse covariance $B(t)$ in \eqref{eq:B}
\begin{lemma}
(Matrix Azuma Inequality~\cite{tropp2012user}) Consider the sequence $\{M_k\}_{1\leq k\leq K}$ of symmetric $d \times d$ random matrices adapted to some filtration $\{\mathscr{G}_k\}_{1\leq k\leq K}$, such that $\mathbb{E}[M_k|\mathscr{G}_{k-1}] = 0$. Assume that there is a deterministic sequence of symmetric matrices $\{A_k\}_{1\leq k\leq K}$ that satisfy $M_k^2 \preceq A_k^2$ , almost surely. Let $\sigma^2 = \|\sum_{1\leq k\leq K} A_k^2\|$. Then, for all $\varepsilon\geq 0$, it holds that
\begin{flalign}
\mathbb{P}\left(\lambda_{\max} \left(\sum_{k=1}^K M_k\right) \geq \varepsilon\right) \leq d \cdot e^{-\varepsilon^2/8\sigma^2}.\nonumber
\end{flalign}
\label{lem:3}
\end{lemma}
Lemma \ref{lem:4} provides a high probability lower bound for the minimum eigenvalue of $B(t)$. Then, Lemma \ref{lem:5} bounds the estimation error. 
\begin{lemma}
For $B(t)$ in \eqref{eq:B} and $t\leq T$, on the event $W_T$ defined in \eqref{eq:WT}, by Lemma 2 and 3, with the probability at least  $1-\delta$, we have
\label{lem:4}
\begin{flalign}
\lambda_{\min} (B(t)) \geq \lambda_{s1}(t-1) \left(1 - \sqrt{\frac{32v_T(\delta)^4}{t-1}\log\frac{d_y T}{\delta}} \right).\nonumber
\end{flalign}
\end{lemma}
\begin{lemma}
In Algorithm 1, let $\widehat{\eta}(t)$ be the parameter estimate, as defined in \eqref{eq:etahat}. Then, for $t\leq T$, on the event $W_T$ defined \eqref{eq:WT}, we have
\begin{eqnarray}
\mathbb{P}\left(\|\widehat{\eta}(t)-\eta_*\| > \varepsilon |B(t) \right) \leq  2  e^{-\frac{\varepsilon^2}{2d_y \lambda_{\max}(B(t)^{-1})\gamma^2_{ry}}}.\nonumber
\end{eqnarray}
\label{lem:5}
\end{lemma}
Next, Lemma \ref{lem:6} gives an upper bound for the probability that Algorithm \ref{algo} does not choose the optimal arm at time $t$. Finally, Lemma \ref{lem:7} studies the weighted sum of indicator functions $I(a^*(t) \neq a(t))$ that count the effective number of times that the algorithm chooses sub-optimal arms.
\begin{lemma}
Given $B(t)$, an upper bound of probability of choosing a sub-optimal arm is bounded as follows:
\begin{flalign}
\mathbb{P}(a^*(t)\neq a(t)|B(t)) \leq \frac{
 2 N\lambda_{s2}^{1/2}d_y v_T(\delta) \gamma_{ry}}{\sqrt{\eta_*^T S_y \eta_* }}\lambda_{t}^{1/2}, \nonumber
\end{flalign}
where $\lambda_t = \lambda_{\max}(B(t)^{-1})$. 
\label{lem:6}
\end{lemma}
\begin{lemma}
For $I(a^*(t) \neq a(t))$, on the event $W_T$, with the probability at least $1-\delta$, we have 
\begin{flalign}
\sum_{t^* \leq t \leq T } \frac{1}{\sqrt{t-1}} I(a^*(t) \neq a(t))\leq \sqrt{32\log T \log (T\delta^{-1})}
+ \sum_{t^* \leq t \leq T} \frac{1}{\sqrt{t-1}} \mathbb{P}(a^*(\tau) \neq a(\tau)|B(t)),\nonumber
\end{flalign}
where $t^* = 128 v_T(\delta)^4 \log \frac{d_y T}{\delta} + 1$.
\label{lem:7}
\end{lemma}
Note that $\mathrm{Regret}(T)$ is the sum of the conditional expected reward difference $\left(y_{a^*(t)}(t) - y_{a(t)}(t)\right)^\top \eta_*$ for $1\leq t \leq T$. The difference $\left(y_{a^*(t)}(t) - y_{a(t)}(t)\right)^\top \eta_*$ at time $t$ is greater than 0, only when $a^*(t)\neq a(t)$. Thus, the regret can be rewritten as $\mathrm{Regret}(T) = \sum_{t=1}^T \left(y_{a^*(t)}(t) - y_{a(t)}(t)\right)^\top \eta_*I(a^*(t)\neq a(t))$. To find an upper bound of the regret, we find high probability upper bounds for $\left(y_{a^*(t)}(t) - y_{a(t)}(t)\right)^\top \eta_*$ and $I(a^*(t)\neq a(t))$, respectively. For both upper bounds, the inverse of the (unnormalized) empirical covariance matrix $B(t)$ in \eqref{eq:B} matters in that the matrix determines the size of estimation error $\|\widehat{\eta}(t)-\eta_*\|$. 

By, Lemma \ref{lem:4}, we have
\begin{eqnarray}
\lambda_{\min} (B(t)) \geq \lambda_{s1}(t-1) \left(1 - \sqrt{\frac{32v_T(\delta)^4}{t-1}\log\frac{d_y T}{\delta}} \right),\label{eq:eig2}
\end{eqnarray}
for all $1 \leq t \leq T$ with the probability at least $1-2\delta$. This implies that $B(t)$ grows linearly with the horizon almost surely.
Next, we investigate the estimation error $\|\eta_* - \widehat{\eta}(t)\|$ based on the above result of the minimum eigenvalue of $B(t)$. Using $\|y_i(t)\|_{\infty} \leq \lambda_{s2}^{1/2} v_{T}(\delta)$ on the event $W_T$, we have
\begin{flalign}
(y_{a^*(t)}(t) - y_{a(t)}(t) )^\top \eta_* 
 \leq \lambda_{s2}^{1/2} v_{T}(\delta) \| \widehat{\eta}(t)-\eta_*\|,\label{eq:bderror}
\end{flalign}
where $\lambda_{s2} = \lambda_{\max}(S_y) $.
So, we write the regret in the following form:
\begin{eqnarray}
\mathrm{Regret}(T)
\leq \sum_{t=1}^T \lambda_{s2}^{1/2} v_{T}(\delta) \| \widehat{\eta}(t)-\eta_*\| I(a^*(t) \neq a(t)).\label{eq:reg2}
\end{eqnarray}	
Here, we denote
$\lambda_t =  \lambda_{\max}(B(t)^{-1}) = (\lambda_{\min}(B(t)))^{-1}$. By \eqref{eq:eig2}, we can find $t^* = 128 v_T(\delta)^4 \log \frac{d_y T}{\delta} + 1$, such that 
\begin{eqnarray}
 \lambda_{t}  \leq \frac{2}{\lambda_{s1} (t-1)}, \label{eq:lambdat}
\end{eqnarray}
with the probability at least $1-\delta$, for all $t^* < t \leq T$. By Lemma \ref{lem:5} and \eqref{eq:lambdat}, for all $t^* < t \leq T$, 
with the probability at least $1-3\delta$, we have
\begin{flalign}
\lambda_{s2}^{1/2} v_{T}(\delta) \| \widehat{\eta}(t)-\eta_*\| \leq a_1(t-1)^{-1/2},~~~~~~\label{eq:bderror2}
\end{flalign}
where $a_1 = 4(\lambda_{s2}/\lambda_{s1})^{1/2} v_T(\delta) \sqrt{2d_y \log(2T\delta^{-1})}$. Thus, with $(y_{a^*(t)} -y_{a(t)})^\top \eta_* \leq 2\lambda_{s2}^{1/2}v_{T}(\delta) \|\eta_*\|$ for $t<t^*$, the regret can be represented
\begin{flalign}
\mathrm{Regret}(T) &\leq  \sum_{ t < t^*}  2\lambda_{s2}^{1/2}v_{T}(\delta) \|\eta_*\|\nonumber\\
&+ \sum_{t^* \leq t \leq T} a_1(t-1)^{-1/2} I(a^*(t) \neq a(t)),
\end{flalign}
with the probability at least $1-3\delta$. Now, we consider the probability to choose the optimal arm at time $t$. By Lemma \ref{lem:6}, we have 
\begin{flalign}
\sum_{t^* \leq t \leq T} \frac{\mathbb{P}(a^*(t)\neq a(t)|B(t))}{\sqrt{t-1}} 
 \leq \frac{ 2^{3/2} N \lambda_{s2}^{1/2} d_y v_T(\delta) \gamma_{ry} }{ \|\eta_*\|\lambda_{s1}^{1/2} } \log T.\label{eq:indi2}
\end{flalign} 
Now, we construct an upper bound about the indicator function $I(a^*(t)\neq a(t))$ in \eqref{eq:reg2}, by Lemma \ref{lem:7}.
\begin{flalign}
\sum_{t^* \leq t \leq T } \frac{1}{\sqrt{t-1}} I(a^*(t) \neq a(t))\leq \sqrt{32\log T \log (T\delta^{-1}) }+ \sum_{t^* \leq t \leq T} \frac{1}{\sqrt{t-1}} \mathbb{P}(a^*(\tau) \neq a(\tau)|B(\tau)),\label{eq:indi}
\end{flalign}
with the probability at least $1-\delta$. Therefore, by  \eqref{eq:indi2} and \eqref{eq:indi}, with the probability at least $1-4\delta$, the following inequalities hold for the regret of the algorithm, which yield to the desired result:
\begin{eqnarray}
\mathrm{Regret}(T)&=&\sum_{t=1}^T (y_{a^*(t)}(t)-y_{a(t)}(t))^\top \eta_* I(a^*(t) \neq a(t))\nonumber\\
&\leq& 2\lambda_{s2}^{1/2}v_T(\delta)\|\eta_*\| t^*+ \sum_{t^* \leq t \leq T} a_1\frac{1}{\sqrt{t-1}} I(a^*(t) \neq a(t)) \nonumber\\
&=& O\left( \frac{\lambda_{s2}}{\lambda_{s1}}\gamma_{ry} Nd_y^{2/3} \left(\log \frac{Nd_y T}{\delta}\right)^{5/2} \log \frac{d_y T}{\delta}  \right).~~~
\end{eqnarray}
Finally, using $S_y = A\Sigma_xA^\top + \Sigma_y$, $\lambda_{s2}/\lambda_{s1} = O((\lambda_{a2}+\lambda_{y2})/(\lambda_{a1}+\lambda_{y1}))$, with the probability at least $1-4\delta$, we have
\begin{flalign}
\mathrm{Regret}(T)=O\left( \frac{(\lambda_{a2}+\lambda_{y2})\gamma_{ry}}{\lambda_{a1}+\lambda_{y1}} Nd_y^{3/2} \left(\log \frac{Nd_y T}{\delta}\right)^{5/2} \log \frac{d_y T}{\delta}  \right).
\end{flalign}
This bound is relatively looser in terms of $N,~d_y$ and tighter in terms of $T$ as compared to the bound $O(polylog(N)\sqrt{d_xT})$ for fully observable contexts \cite{chu2011contextual,agrawal2013thompson}. But, this looser bound in terms of $N,~d_y$ is created to improve the regret bound in terms of $T$. 
\end{proof}

\section{Numerical Illustrations}
\label{sec:5}

In this section, we perform numerical analyses for the theoretical result in the previous section. We simulate cases for $N=10,20,50$ and different dimensions of the observations $d_y=5,20,50$ with a fixed context dimension $d_x=20$. Each case is repeated $100$ times and the average and worst quantities of $100$ scenarios are reported. 

\begin{figure}[ht]
    \centering
    \psfrag{N=10~~~~}{\scriptsize$N=10$}
    \psfrag{N=20~~~~}{\scriptsize$N=20$}
    \psfrag{N=50~~~~}{\scriptsize$N=50$}
    \psfrag{N=100~~~~}{\scriptsize$N=100$}
    \psfrag{dy=5~~~~}{\scriptsize$d_y=5$}
    \psfrag{dy=10~~~~}{\scriptsize$d_y=10$}
    \psfrag{dy=20~~~~}{\scriptsize$d_y=20$}
    \psfrag{dy=50~~~~}{\scriptsize$d_y=50$}
    \psfrag{N~~~~~~~~~~~~~~~~~~~~~~}{~~~~~~~~~~Normalized Regret vs Time}
    \psfrag{N~~~~~~~~~~~~~~~~~~~~~~~~}{~~~~~~~~~~~~Normalized Regret vs Time}
    \psfrag{t}{$t$}
    \includegraphics[width=0.45\textwidth]{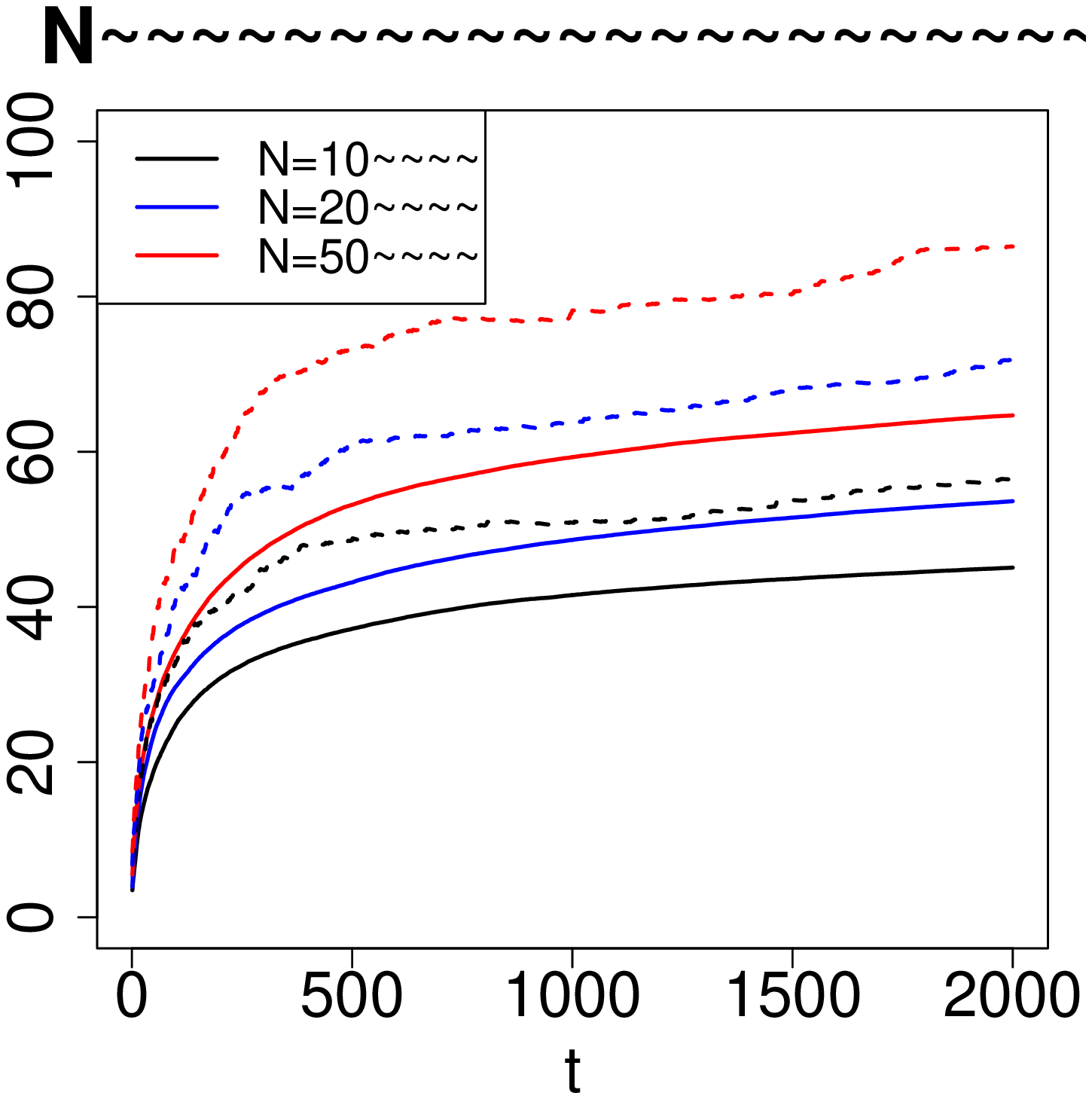}
    \includegraphics[width=0.45\textwidth]{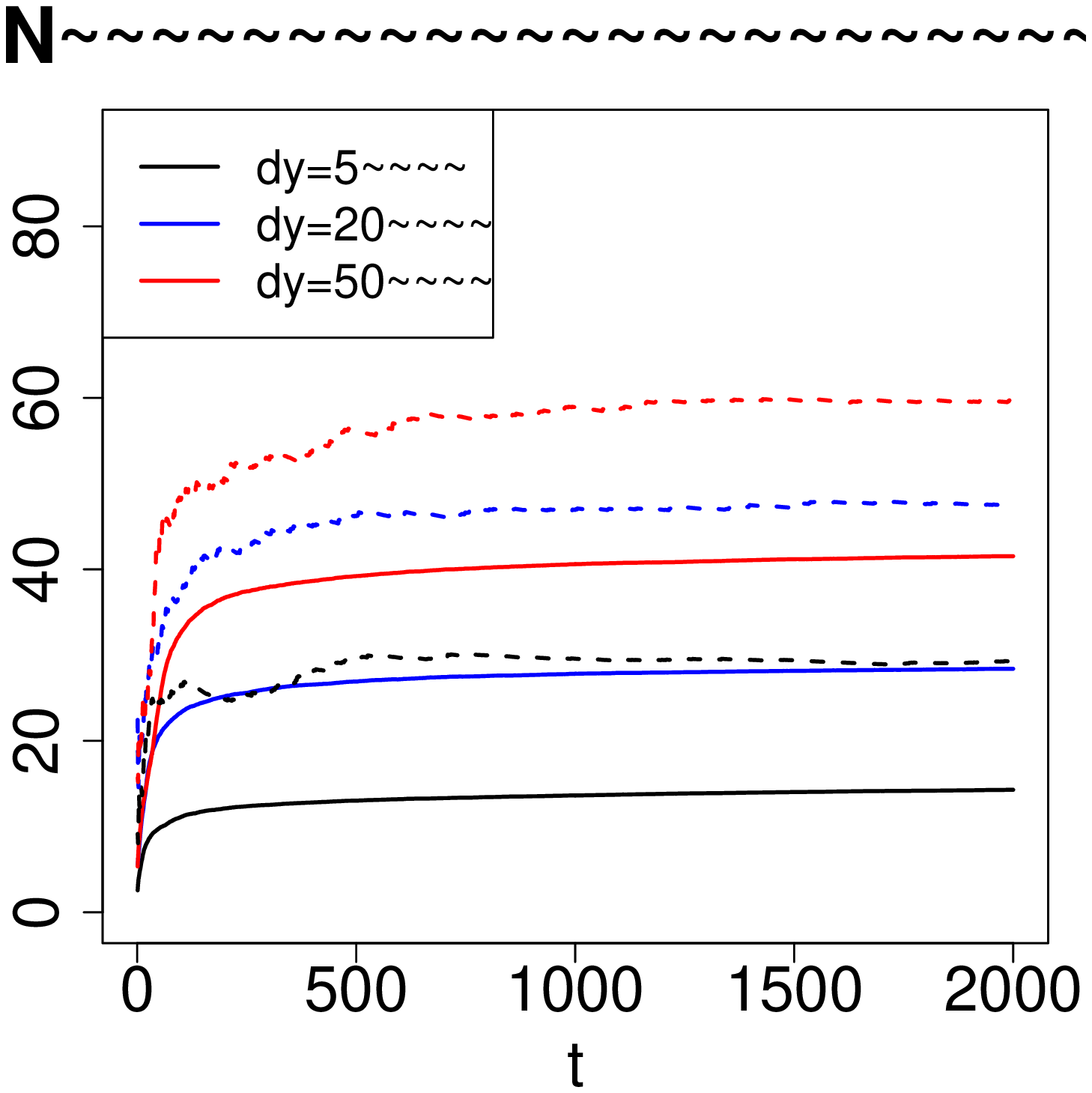}
    \normalsize 
    \caption{Plots of $\mathrm{Regret}(t)/\log t$ over time for the different number of arms $N = 10,20,100$ and  $d_y=5,20,50$. The solid and dashed lines represent average and worst regret curves, respectively.}
    \label{fig:1}
\end{figure}

\begin{figure}[h]
    \centering
    \psfrag{N=10~~~~}{\scriptsize$N=10$}
    \psfrag{N=20~~~~}{\scriptsize$N=20$}
    \psfrag{N=50~~~~}{\scriptsize$N=50$}
    \psfrag{N=100~~~~}{\scriptsize$N=100$}
    \psfrag{dy}{$d_y$}
    \psfrag{A~~~~~~}{\scriptsize Average}
    \psfrag{90\%}{\scriptsize 90\%}
    \psfrag{W~~~~}{\scriptsize Worst}
    \psfrag{N~~~~~~~~~~~~~~~~~~~~~~~~}{~~~~~~~~~~~~~~~~~~~~Regret at $T=2000$}
    \includegraphics[width=0.8\textwidth]{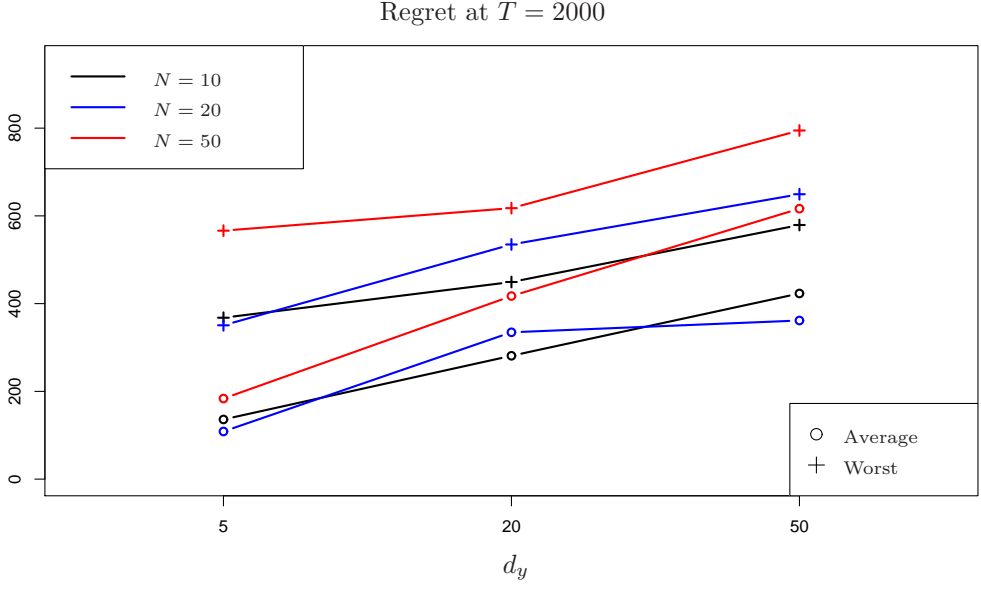}
    \normalsize
    \caption{Plot of average and worst-case $\mathrm{Regret}(T)$ at $T=2000$ for different number of arms $N = 10,20,50$ and dimension of observations $d_y = 5,20,50$. }
    \label{fig:2}
\end{figure}

For Figure \ref{fig:1}, the left plot depicts the average (solid) and worst-case (dashed) regret among all scenarios, normalized by $\log t$. The number of arms $N$ varies as shown in the graph, while the dimension is fixed to $d_y=10$. Next, the right one illustrates that the normalized regrets increase over time for different $d_y$ at the fixed number of arms $N=5$. For both plots, the worst-case regret curves are well above the average ones, but the slopes of curves for both cases become flat as time goes on, implying that the worst-case regret grows logarithmically in terms of $t$ as well. Figure \ref{fig:2} presents the average and worst-case regret (non-normalized) at time $T=2000$ for different $N=10,20,50$ and $d_y=5,20,50$. The plot shows that the regret at $T=2000$ increase as $N$ and $d_y$ become larger. In addition, it shows that the dimension of observations $d_y$ has a greater effect on the regret than that of the number of arms $N$.


\section{Conclusion}
\label{sec:6}

This work investigates reinforcement learning algorithms for contextual bandits where the contexts are observed imperfectly focusing on the theoretical results about the regret growth. We establish a high probability regret bound for Greedy algorithms, which grows poly-logarithmically with the horizon $T$.

There are multiple interesting future directions introduced in this paper. First, it will be of interest to study reinforcement learning policies for settings that each arm has its own parameter. Further, regret analysis for contextual bandits under imperfect context observations where the other parameters such as the covariance matrices of contexts and observations and the sensing matrix are unknown, is another problem for future work.

\section{Appendices}

\subsection{Proof of Lemma 1}
Note that $S_y^{-0.5} y_i(t)$ has the normal distribution $N(0,I_{d_y})$. Then, we have
\begin{eqnarray}
\mathbb{P}\left( |y_{ij}(t)| \geq \varepsilon \right) \leq 2 \cdot e^{-\frac{\varepsilon^2}{2}}
\end{eqnarray}
where $y_{ij}(t)$ is the $j$th component of $y_{i}(t)$. By plugging $v_T(\delta)$ to $\varepsilon$, we have
\begin{eqnarray}
\mathbb{P}\left( |y_{ij}(t)| \geq v_T(\delta) \right) \leq 2 \cdot e^{-\frac{v_T(\delta)^2}{2}} = 2 \cdot e^{-\log\frac{2Nd_yT}{\delta}} = \frac{\delta}{Nd_yT}. 
\end{eqnarray}
Thus,
\begin{eqnarray}
\mathbb{P} (W_T) \geq 1 - \sum_{t=1}^T\sum_{i=1}^N\sum_{j=1}^{d_y} \mathbb{P}\left( |y_{ij}(t)| \geq v_T(\delta) \right) \geq 1-\delta.
\end{eqnarray}

\subsection{Proof of Lemma 2}
We use the following decomposition 

\begin{eqnarray}
S_y^{-0.5} y_{a(t)}(t) = P_{C(S_y^{0.5}\widehat{\eta}(t))} S_y^{-0.5} y_{a(t)} (t) + P_{C(S_y^{0.5}\widehat{\eta}(t))^\perp} S_y^{-0.5} y_{a(t)} (t). \label{eq:decomp}
\end{eqnarray}

We claim that $P_{C(S_y^{0.5}\widehat{\eta}(t))} S_y^{-0.5} y_{a(t)}$ and $P_{C(S_y^{0.5}\widehat{\eta}(t))^\perp} S_y^{-0.5} y_{i}(t) $ are statistically independent. To show it, define
\begin{eqnarray}
Z(\nu,N) = \underset{Z_i,1 \leq i \leq N}{\argmax}\left\{ Z_i^\top \nu \right\}\label{eq:zmun},
\end{eqnarray}
where $Z_i$ has the distribution $N(\mathbf{0}_{d_y},I_{d_y})$ and $\nu$ is an arbitrary vector in $\mathbb{R}^{d_y}$. The vector $Z_i$ can be decomposed as $Z_i = P_{C(\nu)} Z_i + (I_d-P_{C(\nu)}) Z_i$.
Then, we have $Z(\nu,N) = \underset{Z_i,1 \leq i \leq N}{\argmax} \left\{ (P_{C(\nu)} Z_i)^\top \nu  \right\}$, because $P_{C(\nu)}\nu = \nu$. This implies that only the first term of the decomposed terms, $P_{C(\nu)} Z_i$, affects the result of $\underset{Z_i,1 \leq i \leq N}{\argmax} \left\{ Z_i^\top  \nu  \right\}$. This means that $Z(\nu,N)$ has the same distribution as $P_{C(\nu)} Z(\nu,N) +(I_d- P_{C(\nu)})Z_{i}$, which means
\begin{eqnarray}
Z(\nu,N) \overset{d}{=}P_{C(\nu)} Z(\nu,N) +(I_d- P_{C(\nu)})Z_{i}\label{eq:equaldis},
\end{eqnarray}
where $\overset{d}{=}$ is used to denote the equality of the probability distributions. Note that 
$$ S_y^{-0.5} y_{a(t)} = \argmax_{S_y^{-0.5} y_{i}, 1\leq i \leq N} (S_y^{-0.5} y_{i}(t))^\top S_y^{0.5}\widehat{\eta}(t). $$ Thus, $S_y^{-0.5} y_{a(t)}$ has the same distribution as $P_{C(S_y^{0.5}\widehat{\eta}(t))} S_y^{-0.5} y_{a(t)}  + P_{C(S_y^{0.5}\widehat{\eta}(t))^\perp} S_y^{-0.5} y_{i}(t) $, where $P_{C(S_y^{0.5}\widehat{\eta}(t))} S_y^{-0.5} y_{a(t)}$  and $P_{C(S_y^{0.5}\widehat{\eta}(t))^\perp} S_y^{-0.5} y_{i}(t) $ are statistically independent. By the decomposition \eqref{eq:decomp} and the independence, \\
$\mathbb{E}\left[ S_y^{-0.5}y_{a(t)}(t)y_{a(t)}(t)^\top S_y^{-0.5}  | \mathscr{F}_{t-1} \right]$  can be written as

\begin{flalign}
&\mathbb{E}\left[ S_y^{-0.5}y_{a(t)}(t)y_{a(t)}(t)^\top S_y^{-0.5}  | \mathscr{F}_{t-1} \right] \nonumber\\
&= \mathbb{E}\left[ (P_{C(S_y^{0.5} \widehat{\eta}(t))} + P_{C(S_y^{0.5} \widehat{\eta}(t))^\perp}) S_y^{-0.5}y_{a(t)}(t)y_{a(t)}(t)^\top S_y^{-0.5} (P_{S_y^{0.5} \widehat{\eta}(t)} + P_{S_y^{0.5} \widehat{\eta}(t)^\perp}) | \mathscr{F}_{t-1} \right]\nonumber\\
&= \mathbb{E}\left[ P_{C(S_y^{0.5} \widehat{\eta}(t))} S_y^{-0.5}y_{a(t)}(t)y_{a(t)}(t)^\top S_y^{-0.5} P_{C(S_y^{0.5} \widehat{\eta}(t))} | \mathscr{F}_{t-1} \right]  +  P_{C(S_y^{0.5} \widehat{\eta}(t))^\perp} .
\end{flalign}

To proceed, we show that the first term above, $\mathbb{E}[ P_{S_y^{0.5}\widehat{\eta}(t)}  S_y^{-0.5} y_{a(t)}(t)y_{a(t)}(t)^{\top} S_y^{-0.5}P_{S_y^{0.5}\widehat{\eta}(t)} |\widehat{\eta}(t)] =a P_{S_y^{0.5}\widehat{\eta}(t) }$  for some constant $a>1$. Using $P_{C(\nu)} = \nu\nu^\top/\nu^\top \nu$ for an arbitrary vector $\nu \in \mathbb{R}^{d_y}$, we have
\begin{eqnarray}
&& P_{S_y^{0.5}\widehat{\eta}(t) } \mathbb{E}[ S_y^{-0.5} y_{a(t)}(t) y_{a(t)}(t)^{\top} S_y^{-0.5}|\widehat{\eta}(t) ]P_{S_y^{0.5}\widehat{\eta}(t)}\nonumber\\
&=& \frac{S_y^{0.5}\widehat{\eta}(t)\widehat{\eta}(t)^{\top}S_y^{0.5} }{\widehat{\eta}(t)^{\top}S_y \widehat{\eta}(t)} \mathbb{E}[ S_y^{-0.5} y_{a(t)}(t)y_{a(t)}(t)^{\top} S_y^{-0.5}|\widehat{\eta}(t) ] \frac{S_y^{0.5}\widehat{\eta}(t)\widehat{\eta}(t)^{\top}S_y^{0.5} }{\widehat{\eta}(t)^{\top}S_y \widehat{\eta}(t)} \nonumber\\
&=& \frac{S_y^{0.5}\widehat{\eta}(t)}{\widehat{\eta}(t)^{\top} S_y \widehat{\eta}(t)} \mathbb{E}[ (\widehat{\eta}(t)^{\top}  S_y^{0.5} S^{-0.5}_y y_{a(t)}(t))^2|\widehat{\eta}(t) ]\frac{\widehat{\eta}(t)^{\top}S_y^{0.5}}{\widehat{\eta}(t)^{\top}S_y\widehat{\eta}(t)} \nonumber\\
&=& P_{S_y^{0.5}\widehat{\eta}(t) } \mathbb{E}\left[ \left. \left( \left(\overrightarrow{S_y^{0.5}\widehat{\eta}(t)}\right)^\top S_y^{-0.5} y_{a(t)}(t)   \right)^2\right|\widehat{\eta}(t) \right],
\end{eqnarray}

where $\overrightarrow{S_y^{0.5}\widehat{\eta}(t)} = S_y^{0.5}\widehat{\eta}(t)/\|S_y^{0.5}\widehat{\eta}(t))\|$ is the unit vector aligned linearly with $S_y^{0.5}\widehat{\eta}(t)$. Now, it suffices to prove that $$ \mathbb{E}\left[ \left. \left( \left(\overrightarrow{S_y^{0.5}\widehat{\eta}(t)}\right)^\top \left(S_y^{-0.5} y_{a(t)}(t)\right)   \right)^2\right|\widehat{\eta}(t) \right]>1.$$ Note that $\left(\overrightarrow{S^{0.5}_y\widehat{\eta}(t)}\right)^\top S^{-0.5}_y y_{i}(t)$ has the standard normal distribution, since $S^{-0.5}_y y_{i}(t)$ has the distribution $N(0, I_{d_y})$. Thus, $\left(\overrightarrow{S_y^{0.5} \widehat{\eta}(t)}\right)^\top S_y^{-0.5} y_{a(t)}(t)$ is the maximum variable of $N$ variables with the standard normal density. Thus, using $$a(t) = \argmax_{1\leq i\leq N} \{ y_i(t)^\top \widehat{\eta}(t) \} = \argmax_{1\leq i\leq N} \left\{ y_i(t)^\top S_y^{-0.5} \overrightarrow{S_y^{0.5}\widehat{\eta}(t)} \right\},$$ we have

\begin{eqnarray}
y_{a(t)}(t)^\top S_y^{-0.5}\overrightarrow{S_y^{0.5} \widehat{\eta}(t)} \overset{d}{=}  \underset{1 \leq i \leq N}{\max} \{V_i:V_i \sim N(0,1) \}.
\end{eqnarray}

where $\overset{d}{=}$ denotes the equality in terms of distribution. As such, we have

\begin{eqnarray}
 \mathbb{E}\left[\left. \left(y_{a(t)}(t)^\top S_y^{-0.5}\overrightarrow{S_y^{0.5} \widehat{\eta}(t)} \right)^2\right|\widehat{\eta}(t)\right] = \mathbb{E}\left[ \left(\underset{1 \leq i \leq N}{\max} (\{V_i:V_i \sim N(0,1) \}\right)^2 \right]\label{eq:kn}.
\end{eqnarray}

We define the quantity in \eqref{eq:kn} as $k_N$, 
\begin{eqnarray}
k_N = \mathbb{E}\left[ \left(\underset{1 \leq i \leq N}{\max} (\{V_i:V_i \sim N(0,1) \}\right)^2 \right],
\end{eqnarray}
which is greater than 1 for $N\geq 2$ and grows as $N$ gets larger, because $\mathbb{E}[V_i^2] = 1 < \mathbb{E}\left[ \left(\underset{1 \leq i \leq N}{\max} (\{V_i:V_i \sim N(0,1) \}\right)^2 \right]$.
Therefore,
\begin{eqnarray}
\mathbb{E}[ S_y^{-0.5} y_{a(t)}(t)y_{a(t)}(t)^{\top} S_y^{-0.5}|\widehat{\eta}(t) ] = P_{C(S_y^{0.5}\widehat{\eta}(t)) } k_N + P_{C(S_y^{0.5}\widehat{\eta}(t))^\perp} = P_{C(S_y^{0.5}\widehat{\eta}(t))} (k_N-1) + I_{d_y}.\label{eq:matdec}
\end{eqnarray}

\subsection{Proof of Lemma 4}

Consider $V_t=S_y^{-1/2} y_{a(t)}(t)y_{a(t)}(t)^{\top} S_y^{-1/2}$ defined in Lemma \ref{lem:2} to identify the behavior of $B(t)$. By Lemma \ref{lem:2}, the minimum eigenvalue of $\mathbb{E}[ V_t|\mathscr{F}_{t-1}]$ is greater than $1$ for all $t$. Thus, for all $t>0$, it holds that
\begin{eqnarray}
\lambda_{\min}\left(\sum_{\tau=1}^{t-1} \mathbb{E}[V_\tau|\widehat{\eta}(\tau) ]  \right) \geq t-1.\label{eq:mineig}
\end{eqnarray}
Now, we focus on a high probability lower-bound for the smallest eigenvalue of $B(t)$. 
On the event $W_T$, the matrix $v_T^2(\delta)I - V_t$ is positive semidefinite for all $i$ and $t$. Let 
\begin{eqnarray}
X_{\tau} &=& V_\tau - \mathbb{E}[V_\tau|\mathscr{F}_{\tau-1}]\nonumber,\\
Y_\tau &=& \sum_{j=1}^\tau \left(V_j - \mathbb{E}[V_j|\mathscr{F}_{j-1}] \right).
\end{eqnarray}

Then, $X_{\tau}=Y_{\tau}-Y_{\tau-1}$ and $\mathbb{E}\left[ X_{\tau} | \mathscr{F}_{\tau-1} \right] = 0$. Thus, $X_{\tau}$ is a martingale difference sequence. Because $v_T^2(\delta)I - V_t \succeq 0$ for all $t\leq T$, $4v_T^4(\delta)I - X_\tau^2 \succeq 0$, for all $\tau\leq T$, on the event $W_T$. By Lemma \ref{lem:3}, we get

\begin{eqnarray}
\mathbb{P}\left(\lambda_{\min} \left(\sum_{\tau=1}^{t-1} X_{\tau}\right) \leq  (t-1)\varepsilon \right)
\leq d_y \cdot \exp\left(-\frac{(t-1)\varepsilon^2}{32v_T^4(\delta)}\right),
\end{eqnarray}

for $\varepsilon \leq 0$. Now, using $\sum_{\tau=1}^{t-1} X_{\tau} = \sum_{\tau=1}^{t-1}V_\tau - \sum_{\tau=1}^{t-1}\mathbb{E}[V_\tau|\mathscr{F}_{\tau-1}]$, together with
\begin{flalign}
&\lambda_{\min}\left(\sum_{\tau=1}^{t-1}V_\tau-\sum_{\tau=1}^{t-1}\mathbb{E}[V_\tau|\mathscr{F}_{\tau-1}]\right) \nonumber\\
&\leq \lambda_{\min}\left(\sum_{\tau=1}^{t-1}V_\tau\right)-\lambda_{\min}\left( \sum_{\tau=1}^{t-1}\mathbb{E}[V_\tau|\mathscr{F}_{\tau-1}] \right)
\end{flalign}
and \eqref{eq:mineig}, we obtain

\begin{flalign}
&P\left( \lambda_{\min} \left(\sum_{\tau=1}^{t-1}V_\tau \right) 
\leq   (t-1) (1 + \varepsilon) \right) \nonumber\\
&\leq d_y \cdot \exp\left(-\frac{(t-1)\varepsilon^2}{32v_T^4(\delta)}\right),
\end{flalign}

where $-1 \leq \varepsilon \leq 0$ is arbitrary, and we used the fact that $\lambda_{\min}\left(\sum_{\tau=1}^{t-1}V_\tau\right) \geq 0$. 
Indeed, using $\sum_{\tau=1}^{t-1}V_\tau=S_y^{-0.5}B(t)S_y^{-0.5}$, on the event $W_T$ defined in \eqref{eq:WT}, for $-1\leq \varepsilon \leq 0$ we have
\begin{flalign}
\mathbb{P}\left( \lambda_{\min} (B(t)) 
\leq   \lambda_{s1} (t-1) (1 + \varepsilon) \right)  \nonumber\\
\leq~d_y \cdot \exp\left(-\frac{(t-1)\varepsilon^2}{32v_T^4(\delta)}\right),\label{eq:eig}
\end{flalign}
where $\lambda_{s1} = \lambda_{\min}(S_y)$. In other words, by equating $d_y \cdot \exp\left(-(t-1)\varepsilon^2/(32v_T^4(\delta)\right)$ to $\delta/T$, \eqref{eq:eig} can be written as
\begin{eqnarray}
\lambda_{\min} (B(t)) \geq \lambda_{s1}(t-1) \left(1 - \sqrt{\frac{32v_T(\delta)^4}{t-1}\log\frac{d_y T}{\delta}} \right),\label{eq:eig2}
\end{eqnarray}
for all $1 \leq t \leq T$ with the probability at least $1-2\delta$.

\subsection{Proof of Lemma 5}

Note that $\widehat{\eta}(t)$ has the distribution $N\left(\mathbb{E}[\widehat{\eta}(t)|\mathscr{F}_{t-1}], \mathrm{Cov}(\widehat{\eta}(t)|\mathscr{F}_{t-1})\right)$ given the observations up to time $t$, where

\begin{eqnarray}
\mathbb{E}[\widehat{\eta}(t)|\mathscr{F}_{t-1}] &=& B(t)^{-1}  \left(\Sigma^{-1}+\sum_{\tau=1}^{t-1} y_{a(\tau)}(\tau) y_{a(\tau)}(\tau)^\top\right) \eta_* = \eta_*\nonumber\\
\mathrm{Cov}(\widehat{\eta}(t)|\mathscr{F}_{t-1}) &=& B(t)^{-1}\gamma^2_{ry}.
\end{eqnarray}

For $Z\sim N(0, \lambda_{\max}(B(t)^{-1})\gamma^2_{ry} )$, using the Chernoff bound, we get

\begin{eqnarray}
\mathbb{P}\left(\|\widehat{\eta}(t)-\eta_*\| > \varepsilon |B(t) \right) &\leq& \mathbb{P}\left( d_y Z^2 > \varepsilon^2 \right) \nonumber\\
&\leq& 2 \cdot \exp\left(-\frac{\varepsilon^2}{2 d_y \lambda_{\max}(B(t)^{-1})\gamma^2_{ry}}\right),\label{eq:cdfeta}~~~~
\end{eqnarray}
where $\varepsilon \geq 0$. 

\subsection{Proof of Lemma 6}

Let $a^{**}(t)$ be the arm with the second largest expected reward at time $t$ and $\eta_{**}$ be a vector such that $y_{a^{*}(t)}(t)^\top \eta_{**}=  y_{a^{**}(t)}(t)^\top \eta_{**}$ and $\theta(y_{a^{*}(t)}(t) - y_{a^{**}(t)}(t), \eta_*-\eta_{**})=0$, where $\theta(x,y)$ is the angle between two vectors $x$ and $y$. Then,
\begin{eqnarray}
(y_{a^{*}(t)}(t) - y_{a^{**}(t)}(t))^\top \eta_* &=& (y_{a^{*}(t)}(t) - y_{a^{**}(t)}(t))^\top \eta_{**} + (y_{a^{*}(t)}(t) - y_{a^{**}(t)}(t))^\top (\eta_{*}-\eta_{**}) \nonumber\\
&=&\|y_{a^{*}(t)}(t) - y_{a^{**}(t)}(t)\| ~\| \eta_{*}-\eta_{**}\|\cos \theta(y_{a^{*}(t)}(t) - y_{a^{**}(t)}(t), \eta_*-\eta_{**})\nonumber\\
&=&\|y_{a^{*}(t)}(t) - y_{a^{**}(t)}(t)\| ~\| \eta_{*}-\eta_{**}\|.
\end{eqnarray}



If $\|y_{a^{*}(t)}(t) - y_{a^{**}(t)}(t)\|~\|\eta_*-\widehat{\eta}(t)\| \leq (y_{a^{*}(t)}(t) - y_{a^{**}(t)}(t))^\top \eta_*$, we can guarantee $a^*(t)=a(t)$. Thus, the probability not to choose the optimal arm at time $t$ given the observations and $B(t)$ is

\begin{flalign}
\mathbb{P}(a^*(t)\neq a(t)|\{y_i(t)\}_{1\leq i \leq N},B(t))  &= \mathbb{P} \left(\left. \|\widehat{\eta}(t)-\eta_*\| >  \frac{(y_{a^{*}(t)}(t) - y_{a^{**}(t)}(t))^\top \eta_*}{\|y_{a^{*}(t)}(t) - y_{a^{**}(t)}(t)\|} \right|\{y_i(t)\}_{1\leq i \leq N},B(t)\right) \nonumber\\
&\leq 2\cdot\exp\left(-\frac{ \left(\frac{(y_{a^{*}(t)}(t) - y_{a^{**}(t)}(t))^\top \eta_*}{\|y_{a^{*}(t)}(t) - y_{a^{**}(t)}(t)\|}\right)^2}{2d_y\lambda_{\max}(B(t)^{-1})\gamma^2_{ry}}
\right).
\end{flalign}

Using $\|y_{a^{*}(t)}(t) - y_{a^{**}(t)}(t)\|^2 \leq \lambda_{a2} d_y v_T(\delta)^2$ on the event $W_T$,  we have
\begin{eqnarray}
 2\cdot\exp\left(-\frac{ \left(\frac{(y_{a^{*}(t)}(t) - y_{a^{**}(t)}(t))^\top \eta_*}{\|y_{a^{*}(t)}(t) - y_{a^{**}(t)}(t)\|} \right)^2}{2d_y \lambda_{t} \sigma^2_{ry}}\right)
\leq 2\cdot\exp\left(-\frac{ ((y_{a^{*}(t)}(t) - y_{a^{**}(t)}(t))^\top \eta_*)^2}{2d_y^2 v_T(\delta)^2 \lambda_{a2} \lambda_{t} \sigma^2_{ry}}\right).\label{eq:delta}
\end{eqnarray}

 Let $X_1\dots,X_N$ be the order statistics of variables with the standard normal density. The joint distribution of the maximum, $X_N$, and the second maximum variable, $X_{N-1}$, of $N$ independent ones with the standard normal density is

\begin{eqnarray}
f_{X_{(N-1)}, X_{(N)}}(x_{N-1},x_{N})  = N(N-1) \phi(x_N)\phi(x_{N-1}) \Phi(x_{N-1})^{N-2},
\end{eqnarray}

where $\phi$ and $\Phi$ are the pdf and cdf of the standard normal distribution, respectively. The density of $D=X_N-X_{N-1}$, which is the difference of the maximum and second largest variable, can be bounded by $N \phi(0)$ as follows:

\begin{eqnarray}
f_{D}(d) &=& \int f_{D,X_{N-1}}(d, x_{N-1}) d x_{N-1}  \nonumber\\
&=& \int N(N-1) \phi(x_{N-1}+d)\phi(x_{N-1}) \Phi(x_{N-1})^{N-2} dx_{N-1}\nonumber\\
&\leq& N \phi(0).\label{eq:boundforfd}
\end{eqnarray}

Thus, the density $\gamma D$ is bounded by $N\phi(0)/\gamma = N/\sqrt{2\pi\gamma^2}$.

We denote $\Delta_t=(y_{a^{*}(t)}(t) - y_{a^{**}(t)}(t))^\top \eta_*$. The term on the right hand side is the upper bound $\mathbb{P}(a^*(t)\neq a(t)|B(t),\Delta_t)$. Thus, by marginalizing $\Delta_t$ from it, we have
\begin{flalign}
\mathbb{P}(a^*(t)\neq a(t)|B(t)) &=  \int_{-\infty}^{\infty} \mathbb{P}(a^*(t)\neq a(t)|B(t),\Delta_t) f_{\Delta_t}(\Delta_t) d\Delta_t\nonumber\\
&\leq 2 \int_{-\infty}^{\infty} \exp\left(-\frac{ \Delta_t^2}{2d_y^2 \lambda_{a2} v_T(\delta)^2 \lambda_{t} \sigma^2_{ry}}\right) f_{\Delta_t}(\Delta_t) d\Delta_t\nonumber\\
&\leq 2 Nd_y \lambda_{a2}^{1/2}v_T(\delta)\lambda_{t}^{1/2} \gamma_{ry} /\sqrt{\eta_*^T S_y \eta_* }
,\nonumber
\end{flalign} 
where the density of $\Delta_t$,  $f_{\Delta_t}(\Delta_t)$, is bounded by $N/\sqrt{2\pi \eta_*^\top S_y\eta_*}$ by \eqref{eq:boundforfd}.

\subsection{Proof of Lemma 7}

We construct a martingale difference sequence that satisfies the conditions in Lemma \ref{lem:3}. To that end, let $G_1 = H_1 = 0$,
\begin{flalign}
G_\tau = (t-1)^{-1/2} I(a^*(t) \neq a(t)) - (t-1)^{-1/2} \mathbb{P}(a^*(t) \neq a(t)|\mathscr{F}_{t-1}^*),\nonumber
\end{flalign}
and $H_t = \sum_{\tau=1}^t G_\tau$ , where 
\begin{eqnarray}
\mathscr{F}_{t-1}^* = \sigma\{\{B(\tau)\}_{1\leq \tau \leq t-1}\}.\nonumber
\end{eqnarray}
Since $\mathbb{E}[G_\tau|\mathscr{F}_{\tau-1}^*]=0$, the above sequences $\{G_\tau\}_{\tau\geq 0}$ and $\{H_\tau\}_{\tau\geq 0}$ are a  martingale difference sequence and a martingale with respect to the filtration $\{\mathscr{F}_{\tau}^*\}_{1\leq \tau \leq T}$, respectively. Let $c_\tau =2(\tau-1)^{-1/2}$. Since $\sum_{\tau=1}^T |G_\tau| \leq \sum_{\tau=2}^T c_\tau^2 \leq 4\log T$, by Lemma \ref{lem:3}, we have
\begin{flalign}
\mathbb{P}( H_T - H_1 > \varepsilon) \leq  \exp \left(-\frac{\varepsilon^2}{8 \sum_{t=1}^T c_t^2}\right) \leq \exp \left(-\frac{\varepsilon^2}{32\log T }\right).\nonumber
\end{flalign}
Thus, with the probability at least $1-\delta$, it holds that 
\begin{flalign}
\sum_{t^*_T \leq t \leq T } \frac{1}{\sqrt{t-1}} I(a^*(t) \neq a(t)) \leq \sqrt{32\log T \log \delta^{-1} }+ \sum_{t^*_T \leq t \leq T} \frac{1}{\sqrt{t-1}} \mathbb{P}(a^*(\tau) \neq a(\tau)|\mathscr{F}_{\tau-1}^*).\nonumber
\end{flalign}

\ifarxiv
\input{appendix}
\input{Bibliofile}
\else
\bibliographystyle{IEEEtran}   
\bibliography{mybib}      
\fi 
\thispagestyle{empty}

\end{document}